\documentclass[journal]{IEEEtran}

\usepackage{xcolor,soul,framed} 

\colorlet{shadecolor}{yellow}
\usepackage[pdftex]{graphicx}
\graphicspath{{../pdf/}{../jpeg/}}
\DeclareGraphicsExtensions{.pdf,.jpeg,.png}

\usepackage[cmex10]{amsmath}
\usepackage{array}
\usepackage{mdwmath}
\usepackage{mdwtab}
\usepackage{eqparbox}
\usepackage{url}

\begin{document}
\bstctlcite{IEEEexample:BSTcontrol}
    \title{AI-supported Framework of Semi-Automatic Monoplotting for Monocular Oblique Visual Data Analysis}
  \author{Behzad Golparvar, 
      Ruo-Qian Wang \\ 
      \textit{Rutgers University, Department of Civil and Environmental Engineering, Piscataway, NJ, USA}

  \thanks{Email address of corresponding author: rq.wang@rutgers.edu
  }
}


\maketitle

\begin{abstract}
In the last decades, the development of smartphones, drones, aerial patrols, and digital cameras enabled high-quality photographs available to large populations and, thus, provides an opportunity to collect massive data of the nature and society with global coverage. However, the data collected with new photography tools is usually oblique - they are difficult to be georeferenced, and huge amounts of data is often obsolete. Georeferencing oblique imagery data may be solved by a technique called monoplotting, which only requires a single image and Digital Elevation Model (DEM). In traditional monoplotting, a human user has to manually choose a series of ground control point (GCP) pairs in the image and DEM and then determine the extrinsic and intrinsic parameters of the camera to establish a pixel-level correspondence between photos and the DEM to enable the mapping and georeferencing of objects in photos. This traditional method is difficult to scale due to several challenges including the labor-intensive inputs, the need of rich experience to identify well-defined GCPs, and limitations in camera pose estimation. Therefore, existing monoplotting methods are rarely used in analyzing large-scale databases or near-real-time warning systems. In this paper, we propose and demonstrate a novel semi-automatic monoplotting framework that provides pixel-level correspondence between photos and DEMs requiring minimal human interventions. A pipeline of analyses was developed including key point detection in images and DEM rasters, retrieving georeferenced 3D DEM GCPs, regularized gradient-based optimization, pose estimation, ray tracing, and the correspondence identification between image pixels and real world coordinates. Two numerical experiments show that the framework is superior in georeferencing visual data in 3-D coordinates, paving a way toward fully automatic monoplotting methodology.
\end{abstract}


%
\IEEEpeerreviewmaketitle


\section{Introduction}

Historical geography paintings, drawings, and photography are recognized as important data sources for understanding changes on the earth’s surface \cite{piana_topographical_2012,piana_art_2018,watkins_landscapes_2018}. But until the wide application of aerial imagery in the 1980s, the imagery data has not been useful in  quantitative analysis \cite{gabellieri_measuring_2019}. The major reason is that the earlier imagery is often from an oblique perspective and therefore difficult to georeference \cite{gabellieri_measuring_2019,bayr_quantifying_2021}. This georeferencing problem also applies to imagery from emerging technologies such as Unmanned Aerial Vehicles (UAV) and crowdsourcing (including social media and citizen science). The imagery obtained from these data sources, especially those not dedicated to earth surface observation, tend to provide photos and videos at oblique angles. To retrieve valuable data from historical geoscience imagery and harness the power of the widespread new data sources, we need a novel tool to break through the barrier posed by oblique imagery data to enhance the extraction of quantitative geospatial information.

\begin{figure}
    \centering
    \includegraphics[width=\linewidth]{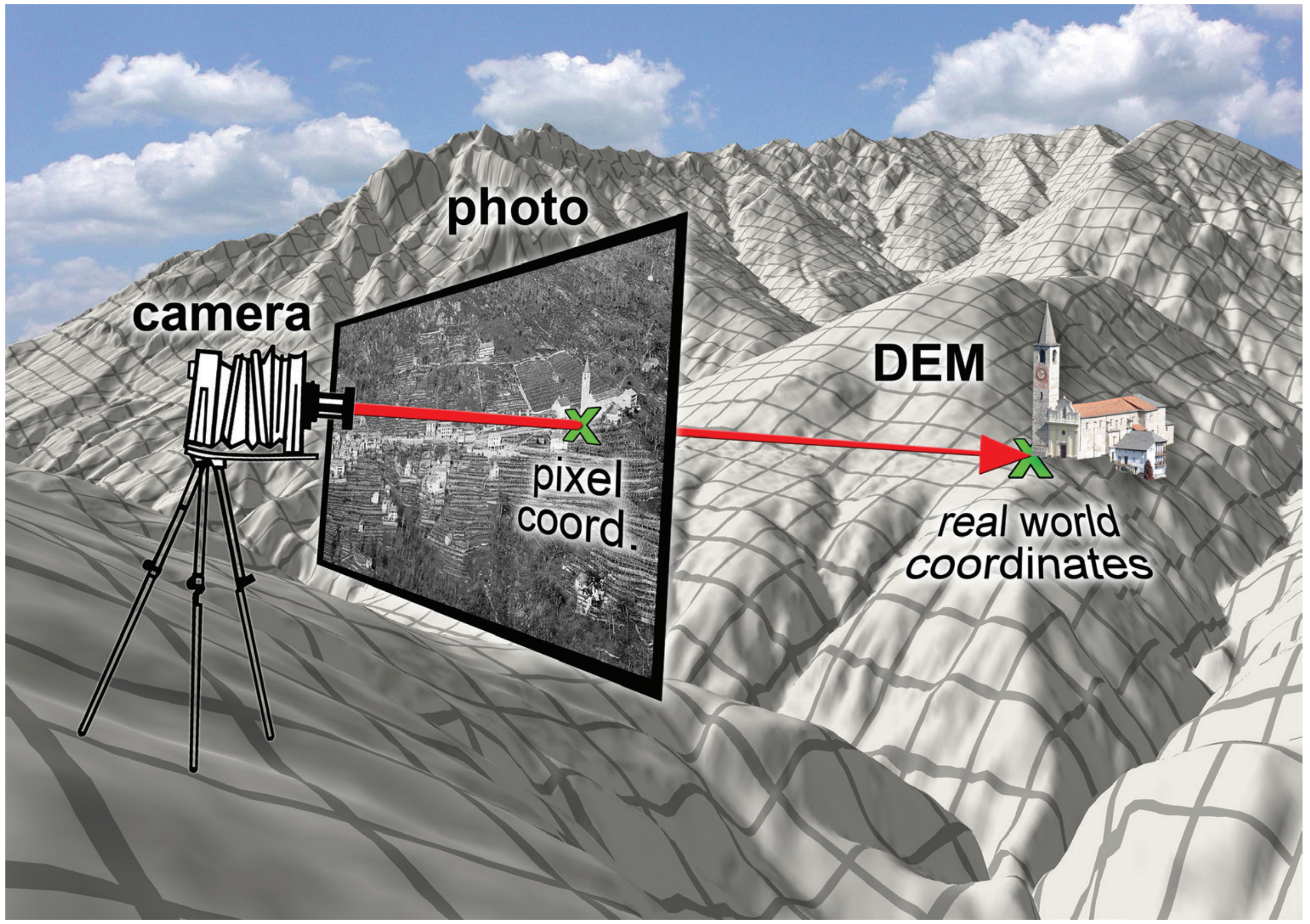}
    \caption{Schematic of the monoplotting technique \cite{gabellieri_measuring_2019}}
    \label{fig:sketch}
\end{figure}

Monoplotting is a potential tool to address the problem, which is a method to obtain correspondence between image pixels and georeferenced 3D space. Originally invented in the 1970s by Makarovi\v{c} \cite{makarovic_digital_1973}, monoplotting was used to transform photographs into georeferenced data (Figure \ref{fig:sketch}) \cite{bayr_quantifying_2021}. The link between the imagery and DEM obtained by the technique of monoplotting has several useful and unique features. First, the link creates pixel-level georeferencing so that the imagery data can be geo-located and mapped in the DEM for further analysis. Second, the final product obtained through monoplotting inherits rich details of the imagery data to reflect the dynamics on the earth surface. Third, in contrast to methods requiring stereo imagery, monoplotting requires only a single image and a DEM to obtain 3-D observation of the scene -- a terrain data-based monocular depth estimation method, which requires low data input to reconstruct 3D information. These unique features of monoplotting allow flexible and powerful applications using imagery data. For example, monoplotting has been used in documenting long-term earth surface changes with repeat photography \cite{pickard_assessing_2002, kull_historical_2005, svenningsen_historical_2015, bayr_automatic_2019}; the ground-based photographs were enabled to supplement aerial photographs with additional information to obtain integrated 3-D view of the scene; because of its low data input of single images, historical photographs from the time before aerial photography can be analyzed quantitatively to provide extraordinary temporal range of earth observation \cite{gimmi_assessing_2016}; the georeferencing feature allows the analysis of crowdsourcing images and suitable to capture the past and emergent events such as flooding \cite{golparvar_ai-enabled_2020, golparvar_ai-supported_2020, triglav-cekada_using_2013}; the oblique data can be used to communicate with the stakeholders leveraging the similarity with daily-life perception and experience of the environment \cite{bozzini_new_2012}; the powerful locating and mapping function enabled low cost solution of field monitoring of slow earth surface dynamics such as glacier movement and vegetation changes \cite{bozzini_new_2012, triglav-cekada_acquisition_2011, triglav_cekada_how_2014, wiesmann_reconstructing_2012}; dynamic information obtained by monoplotting smartphone videos could be used to reconstruct 3-D deformation and movement such as in the dam breaking event for forensic analysis \cite{yuan_unlocking_2020} and multitemporal landslide monitoring \cite{travelletti_correlation_2012}.

Since the pioneering work in the 1970s and 1980s \cite{makarovic_digital_1973, makarovic_data_1983}, efforts have been made to develop software and tools for monoplotting oblique pictures. These include the OP-XFORM project \cite{doytsher_fortran_1995}, the JUKE method \cite{aschenwald_spatio-temporal_2001}, Georeferencing oblique terrestrial photography \cite{corripio_snow_2004}, the 3D Monoplotter \cite{mitishita_3d_2004}, and the DiMoTeP \cite{fluehler_development_2005}. Probably the most applied tool is the WSL Monoplotting Tool (WSL-MPT) developed by the Swiss Federal Research Institute (WSL) to allow the georeferencing and vectorization of topographic photographs. So far, the WSL-MPT has been used for the quantitative analysis of natural hazards \cite{bozzini_new_2012, simao_antunes_do_carmo_using_2018}, glacial processes \cite{wiesmann_reconstructing_2012, scapozza_assessing_2014}, and land cover changes \cite{gabellieri_measuring_2019, stockdale_could_2019, stockdale_could_2019, mccaffrey_assessing_2017}. Varieties of monoplotting tools were also developed in recent years, such as the QGIS plugin Pic2map \cite{milani_pic2map_2014}. None of these products, however, meet the needs of potential end-users in terms of operational flexibility and user-friendly user interface, which greatly inhibit their broad use \cite{bozzini_new_2012}. It should be noted that monoplotting requires an existing high spatial resolution DEM, which are generated for large regions of the Earth surface through other means such as prior stereophotographic mapping or LiDAR. Many high-resolution DEM are currently available through various online databases. Also note that monoplotting is significantly different from photogrammetry: in photogrammetry several images with large overlap are needed, and camera poses are not necessary in all cases. Furthermore, in some cases of aerial photogrammetry an estimate of the camera pose is usually available by GPS measurements, which makes the camera position and parameter calculation much easier. While in monoplotting only one image and the DEM of the region are available and a prior estimate of the camera pose may not be available, which makes the problem more difficult to solve.

\begin{figure}
    \centering
    \includegraphics[width=\linewidth]{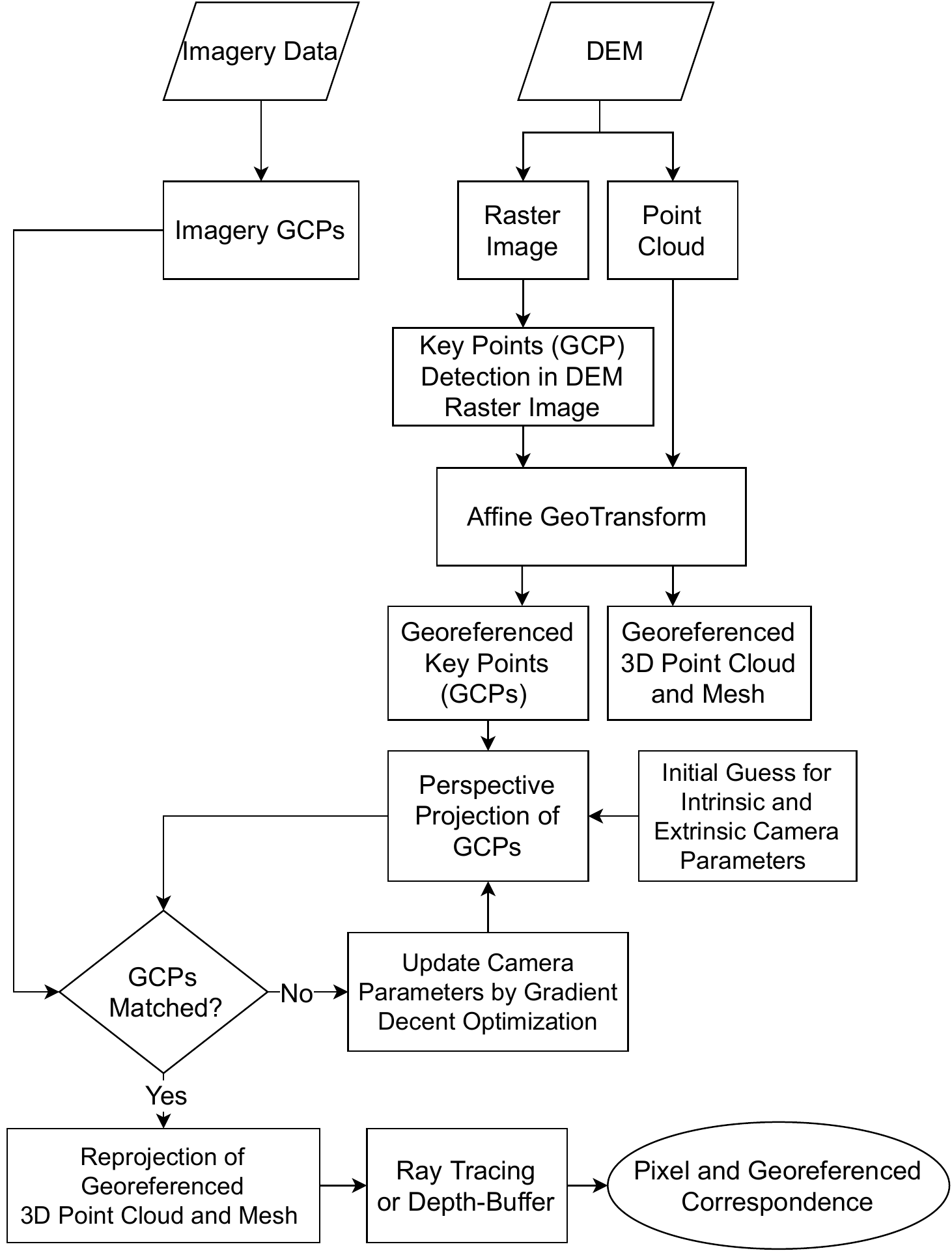}
    \caption{Flowchart of the proposed semi-automatic monoplotting framework.}
    \label{fig:flowchart}
\end{figure}

In the traditional monoplotting method, determining the camera pose requires identifying a series of Ground Control Points (GCP) in the photograph (called image GCPs) and find the corresponding GCPs in DEM of the same area (called DEM GCPs). So the number of GCPs in image and DEM should be equal and the one-to-one correspondence must be established. Then some pose estimation algorithms such as PnP and Directed Linear Transform (DTL) along with Levenberg-Marquardt algorithm \cite{milani_pic2map_2014} must be used to estimate the intrinsic and extrinsic parameters of the camera. For these algorithms, a certain minimum number of GCP pairs is required, e.g. at least 6 pairs of GCPs are required for the DTL algorithm, hence, sometimes users have difficulties to meet such a requirement given the experience and expertise of the method. Furthermore, this method of pose estimation is highly sensitive to positions of GCPs in image and DEM, which users have to manually choose with practical difficulties, due to several technical issues. First, it requires the users to locate the “good” features that can be recognized both in the DEM and imagery data. Hence a process of trial and errors is often required to manually place the GCPs based on the features in both image and DEM raster, subsequently running the pose estimation algorithms and check if the estimated pose makes physically sense. In each attempt user should make sure whether one-to-one relationship of GCPs is correct. Therefore, this makes the process of monoplotting slow and labor intensive. Second, studies showed that good distinguishable GCPs should be uniformly distributed in the image and must not be collinear to enable good results of interactive absolute orientation \cite{triglav_cekada_how_2014}, which sometimes cannot be achieved by naked eyes due to lighting, noise, or background inconsistency. Poor quality of GCPs will result in inaccuracy in the monoplotting results and, sometimes, disable the camera configuration determination caused by code failures or infinite loops. Third, in processing historical images, the landscape may have changed dramatically so that insufficient reliable reference points can be identified on both the historical landscape photograph and a modern DEM \cite{bayr_quantifying_2021}. Fourth, a landscape can be hidden by locating behind buildings, trees, ridges, or back-facing slopes. So, for the GCPs easily identified in the DEM, the user may not find the corresponding ones in the photographs. Finally, the resolution and details of the DEM and photographs will dictate the number and accuracy of the GCPs – higher resolution provides more details of the scenes to recognize features to extract \cite{bayr_quantifying_2021}. Overall, this process requires experience and a large amount of manual work by the user to reach an acceptable accuracy level. Thus, identifying high-quality GCPs is a major bottleneck to scale up monoplotting to process a large volume of data or to design an emergency response/early warning system that requires the capability of real-time processing.

The traditional monoplotting provides useful and unique features to process earth science data but is seriously limited in wide earth science and computer vision applications due to multiple technical challenges. As the imagery data in earth surface observation grows exponentially due to the wide use of smartphones and UAVs, an automatic tool that is scalable for processing a large volume of data is required to meet the challenge of large-scale survey and near-real-time data processing. Specifically, this new tool should meet the following challenges: 1) the GCP identification should be automatic to minimize the human operation burden or at least semi-automatic to guide the user to choose among machine recommendations to significantly reduce the time input; 2) the camera position and configuration determination should be robust and accurate to avoid the difficulty of convergence in optimization; and 3) the data extraction from the imagery data should be automatic or with minimized human training to save time in vectorizing the imagery information.

The objective of this paper is to fill the gap, aiming to develop a reliable and powerful imagery georeferencing and analysis framework. Enabled by computer vision and machine learning techniques, this framework will provide automatic or semiautomatic processing functions to liberate the users from labor-intensive and time-consuming tasks in retrieving pixel-level correspondence between photos and DEM. The method investigated in our study is promising to make monoplotting more useful for wide and quick adoption and large-scale applications, and to allow the broader computer vision, remote sensing, and earth science communities to analyze oblique imagery from historical ground or emerging remotely sensed or crowd-sourcing sources with unprecedented data volume, spatial resolution, and temporal span. To the best of authors' knowledge, this paper is the first time to 1) introduce key point detection into monoplotting, 2) apply regularized gradient-based optimization to significantly improve the accuracy of camera position and parameter determination, and 3) enable the matching of unequal numbers of GCPs from the image data and the DEM. Two numerical experiments with synthetic geometry and real world example showed that the developed framework is reliable and useful, producing satisfactory pixel-level georeferencing results. 

\begin{figure}
    \centering
    \includegraphics[width=\linewidth]{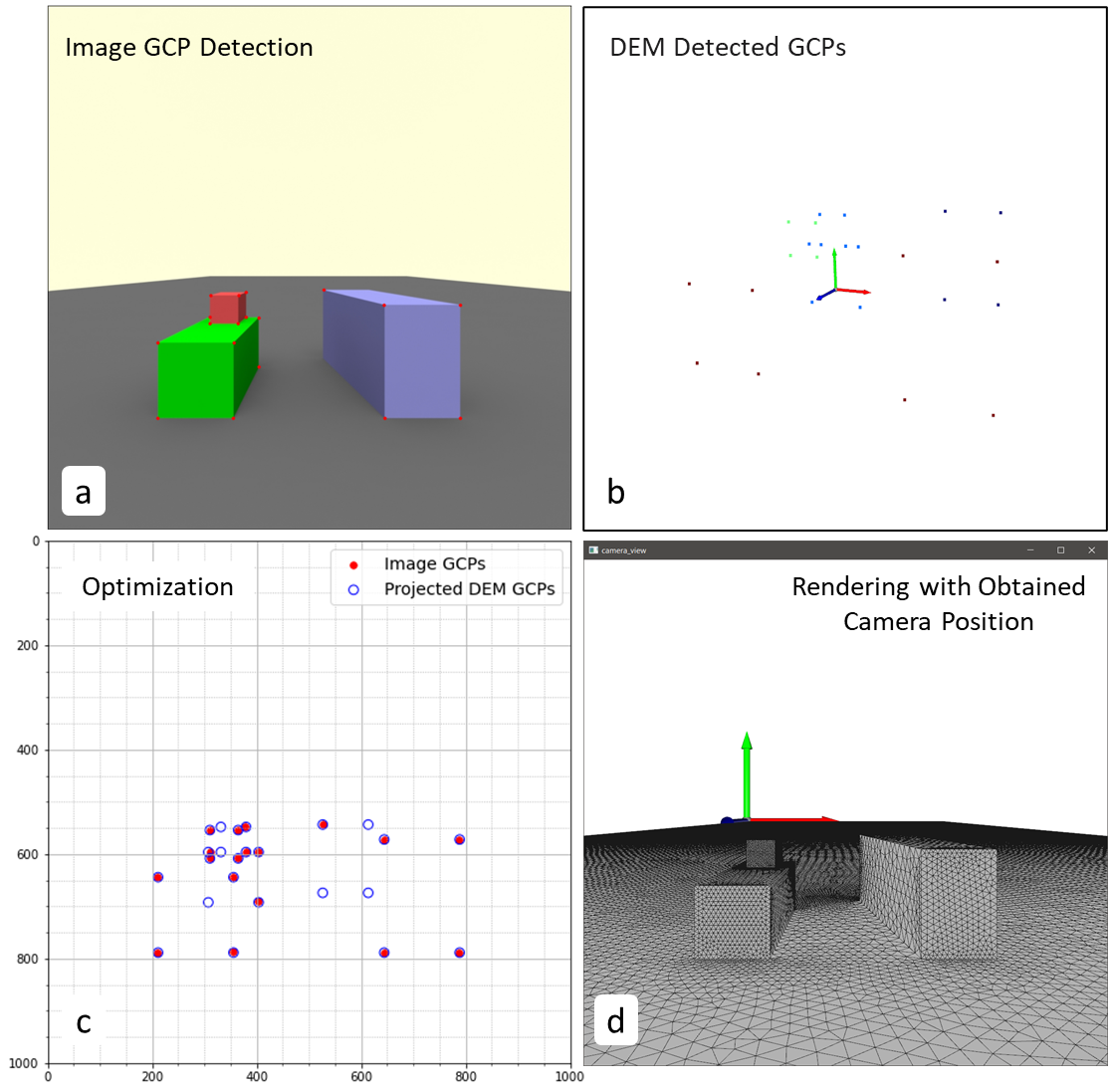}
    \caption{The GCP detection for a synthetic geometry: (a) image GCPs are identified in the synthetic "photo"; (b) DEM GCPs are captured in the elevation "map"; (c) GCP matching on the projected plane; (d) The rendered scene confirms that the obtained camera parameters are correct.}
    \label{fig:3box1}
\end{figure}



\begin{figure*}
    \centering
    \includegraphics[width=\linewidth]{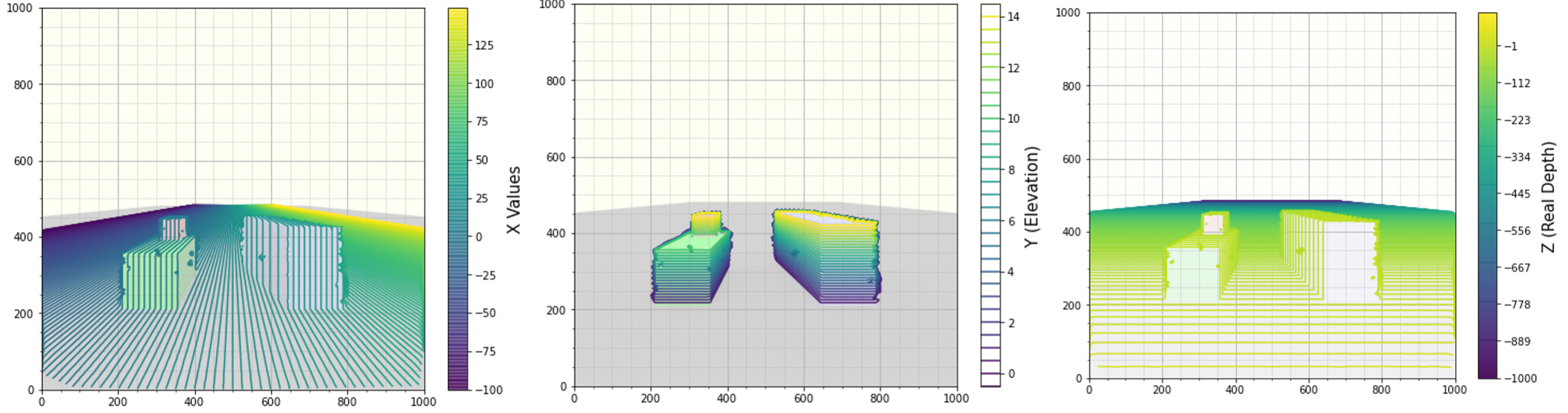}
    \caption{The obtained x, y, and z coordinates: the results confirmed the correct performance of the semi-automatic monoplotting scheme.}
    \label{fig:3box2}
\end{figure*}

\section{Methodology}

The emerging computer vision techniques provide an exciting opportunity to address the issues for wide dissemination and application of the monoplotting technique. In this regard, the GCPs in image data can be detected by employing key point detection algorithms\cite{krizhevsky_imagenet_2017}.  
Several detection algorithms have been developed to extract engineered features, which are mathematical representations of key visual elements in an image such as corners and edges. The general principle behind the algorithms is to identify various kinds of gradients of the grayscale intensity of an image. Popular examples include Harris corner detecter, Shi-Tomasi corner detector, SIFT (Scale-Invariant Feature Transform) \cite{lowe_object_1999}, SURF (Speeded-Up Robust Features) \cite{bay_surf_2006}, and ORB (Oriented FAST and Rotated Brief) \cite{calonder_brief_2012}. In addition, deep learning self-supervised models such as \cite{detone_superpoint_2018} have been proposed for the detection and description of key points in images. Leveraging the progress in this field, we propose to apply a computer-vision-based key point detection algorithm to identify the features in the photographs and DEMs.

In our study, the key point detection method of ORB is employed for detecting 
the GCPs in image and DEM raster. Several benefits and features are enabled by this new algorithm. First, the computer vision algorithm saves time and relaxes the experience requirement of users to manually identify the features from the data. By tuning the sensitivity of the algorithms, an optimal number of GCPs can be obtained for later uses. Second, different numbers of GCPs in the DEM raster and image will be allowed. This feature can benefit the task of GCP identification in multiple ways: 1) the one-to-one corresponding relationship is not necessary to be considered in the identification process so that iterations are, almost, not needed to save the users’ time. 2) Independent GCP identification and image registration provide the critical flexibility for the user to improve the quality of GCPs, e.g., the GCPs can be identified with improved distribution and customized for the lighting and noise level in the image data without worrying to compromise the GCP quality of the DEM raster. 3) The disconnection also enables a new type of interaction between the monoplotting platform and the users – the computer vision scheme can be used as a recommendation system to provide suggestions to the user to choose or modify the GCPs – an efficient and flexible means to ensure the quality of the GCPs. In our framework, shown in Figure \ref{fig:flowchart}, DEM data in the format of a raster image is considered for detecting the GCPs. As mentioned above, the ORB key point detector is used to find the GCPs in the DEM raster. Then, the whole DEM data and detected GCPs that are in pixel coordinate system are transformed into a georeferenced XYZ coordinate system. This is achieved by performing affine geo-transform processing using Eq. \eqref{affine}, where $X_{\rm Geo-UL}$ and $Y_{\rm Geo-UL}$ are the georeferenced position of the upper left pixel in the DEM raster, subscript "p" is for Pixel coordinate system, ${\rm Row}_{\rm rot}$ and ${\rm Col}_{\rm rot}$ are the rotation degrees for cases in which DEM is rotated. In this stage, the georeferenced DEM GCPs are ready for perspective projection to the image plane.

\begin{equation}\label{affine}
\begin{aligned}
    X_{GEO}=X_{\rm Geo-UL}+X_{p}{P}_{\rm width}+Y_{p}{\rm Row}_{\rm rot} \\
    Y_{GEO}=Y_{\rm Geo-UL}+Y_{p}{P}_{\rm height}+X_{p}{\rm Col}_{\rm rot},
\end{aligned}
\end{equation}

\begin{figure*}
    \centering
    \includegraphics[width=\linewidth]{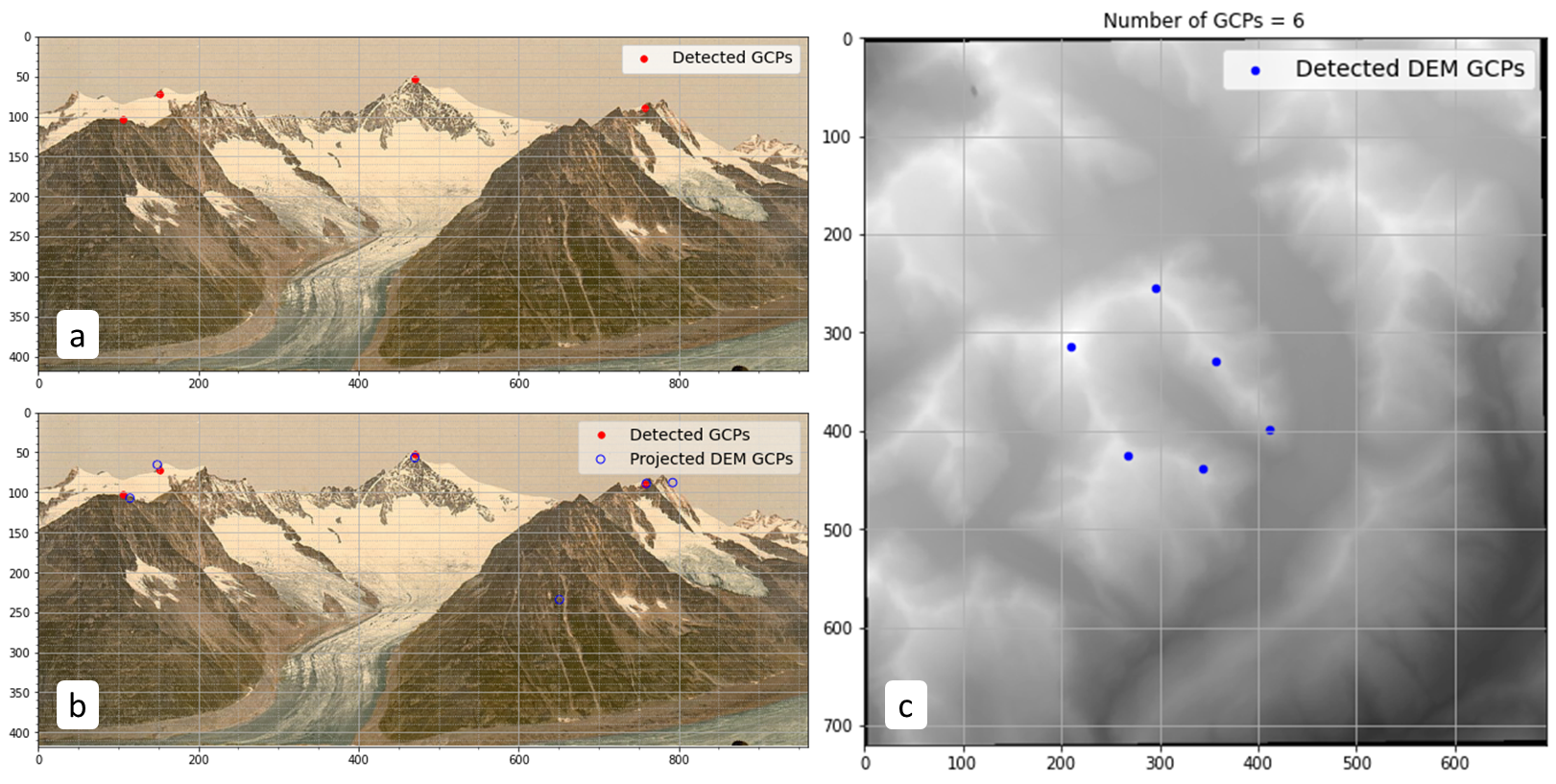}
    \caption{The GCP detection for a photolithograph: (a) image GCPs are identified in the photolithograph; (b) GCP are matched on the projected plane using the regularized optimization scheme; (b) DEM GCPs are captured in the DEM.}
    \label{fig:Glacier1}
\end{figure*}

Next, assuming initial values for seven unknown camera extrinsic and intrinsic parameters including three for position, three for orientation, and the field of view angle, the DEM GCPs are projected to a plane. It should be noted that by having the field of view angle and image width and height, the parameters of focal length and image center can be obtained accordingly. By calculating the Euclidean distance between the projected DEM GCPs with the nearest image GCPs, the loss function is defined. By employing differentiable functions and the method of gradient decent optimization, the camera parameters are iteratively adjusted to reproject the DEM GCPs until the best match between the two sets of GCPs is achieved. Using the camera parameters that enable the best GCP match, the 3-D coordinates of each image pixel are calculated and the objects or events in the imagery data can be mapped in the 3-D space. A flowchart of the monoplotting method is shown in Figure \ref{fig:flowchart} to illustrate the process and the flow of information.

The mismatched number of GCPs in DEM and image poses a difficulty on image registration, which can be addressed by an optimization scheme to match the DEM and photographs to allow the determination of the pixel coordinates in the 3D space. This involves three steps: First, the DEM GCPs are projected to a plane assuming a camera is positioned in a location using the equation below,

\begin{equation}\label{proj}
\begin{matrix}
p=KR\left[P+T\right] \\
     \left[\begin{matrix}u\\v\\1\\\end{matrix}\right]=\left[\begin{matrix}f_x&0&c_x\\0&f_y&c_y\\0&0&1\\\end{matrix}\right]\left[\begin{matrix}R_{11}&R_{12}&R_{13}\\R_{21}&R_{22}&R_{23}\\R_{31}&R_{32}&R_{33}\\\end{matrix}\right]\left[\begin{matrix}X_w+T_x\\Y_w+T_y\\Z_w+T_z\\\end{matrix}\right]
\end{matrix}
\end{equation}\\
where $u$ and $v$ are the horizontal and vertical coordinate of projected points in the image plane, $K$ is the intrinsic camera matrix in which the position of principal point is assumed at the image center, and the focal lengths of $f_{x}$ and $f_{y}$ can be calculated by image dimensions and the field of view angle which is a variable during optimization. $R$ is the rotation matrix and $T$’s components are for the translation vector. Second, the projected GCPs are compared with the GCPs identified in the image GCPs. Finally, the camera parameters and location are iteratively updated using back propagation of functions derivatives to reproject the DEM GCPs until the best comparison is achieved. Since the number of DEM GCPs is different from the image GCPs, we need a new metric to evaluate the goodness of the comparison. This is achieved through a mismatched GCP comparison, in which the distance between the image GCPs and the nearest projected DEM GCP is calculated in the 2D space and the average distance is used to judge the goodness of the comparison.

\begin{equation}\label{loss1}
    S=\frac{\sum_{i=1}^{n}\sqrt{\left(x_{I,i}-x_{DP,nearest}\right)^2+\left(y_{I,i}-y_{DP,nearest}\right)^2}}{n},
\end{equation}
where $x_{I,i}$ and $y_{I,i}$ are the coordinates of the image GCPs in the 2-D space, $x_{DP,nearest}$ and $y_{DP,nearest}$ are the coordinates of the projected DEM GCPs closest to the image GCPs in the 2-D space, $n$ is the number of image GCPs, and $S$ is the average distance to evaluate the goodness of the comparison. This metric allows the user to measure the goodness of comparison between two sets of GCPs in different numbers.

A few studies show that if the GCPs are uniformly distributed in the photograph with good accuracy and precise one-to-one correspondence, the camera position and parameters and the coordinates of the pixels can be determined in high precision, e.g., in the error of 1 m \cite{stockdale_extracting_2015}. Conversely, when the GCPs are suboptimal or when the photos do not exactly correspond to a plane projection of reality (owing to film unflatness, particular lens distortions or other irregularities during photo reproduction or image scanning), the algorithm may converge towards a local minimum that isn’t the best solution, or in the worst cases, the algorithm can even continue to loop without going toward a solution \cite{samtaney_method_1999}. Therefore, a smart and reliable optimization scheme is needed to address the mismatch and the difficulty of convergence and stability.

We propose a machine learning-based optimization scheme to address the issues. A regularization scheme was implemented to promote the computing convergence. In general, the optimization scheme is designed to minimize the following objective function,
\begin{equation}\label{ReglossEq}
    \min_{C}{\ \left(1-\lambda\right)S+\lambda R}
\end{equation}
where $S$ is the average distance defined in Eq. \eqref{loss1}, which is a function of the camera unknown variables ($C$), $\lambda$ is the weight of the regularization, and $R$ is the regularization term.




A preliminary study was conducted to test the new camera configuration determination scheme. We discovered that a consistent and rapid convergence can be achieved using the regularization terms,
\begin{align}
	{\mathit{R}}=\left( \frac{1} {n_{DP}} \sum x_{DP} - \frac {1} {n} \sum x_I \right)^2 \nonumber\\
    +\left( \frac {1} {n_{DP}} \sum y_{DP} - \frac {1} {n} \sum y_I \right)^2,
\end{align}\\
where $n_{DP}$ is the total number of projected DEM GCPs. This regularization term means that the centroid of the projected points should be near the centroid of the image GCPs. This technique accelerates the optimization process and helps reach the best solution with a lower number of iterations. Note that along with this regularization term, other regularization terms can be designed to help optimization converges smoothly in the future. It should be mentioned that the value of $\lambda$ in Eq. \eqref{ReglossEq}, can be determined in the optimization scheme in an adaptive fashion, meaning that when the projected DEM GCPs are far from the image GCPs, $\lambda$ should have larger values, and when convergence is approaching smaller values or zero can be considered.

\section{Experimental Results and Discussion}
Two numerical experiments were conducted by implementing the proposed method using the Python packages of OpenCV \cite{itseez_open_2015}, Open3D \cite{zhou_open3d_2018}, NumPy \cite{harris_array_2020}, and Gdal \cite{gdalogr_contributors_gdalogr_2021}.  PyTorch \cite{paszke_automatic_2017} was used for gradient-based optimization to demonstrate the new semi-automatic monoplotting framework. In the first numerical experiment, a synthetic 3-D geometry consisting of three boxes of different sizes was created and rendered for the test (Figure \ref{fig:3box1}a). The feature detector of ORB was used to identify the key points of the image, i.e., the corners of the rendered boxes. The result shows that after tuning the ORB scheme hyper-parameters, high-quality GCPs can be obtained with a good distribution and accuracy to capture the distinguishing features. Similarly, 24 high-quality DEM GCPs were obtained for the DEM data (i.e. the geometry data in this case) as shown in Figure \ref{fig:3box1}b. The results show that this method is promising to detect the GCPs and saved the user’s great effort in the process, in spite of the idealization of the geometry. 
Figure \ref{fig:3box1}c shows that although the number of the image GCPs is different from the DEM (geometry) GCPs, a good match can still be achieved, indicating a reliable camera parameter determination with flexible GCP numbers. The rendered scene based on the determined camera parameters and the elevation information of the three boxes confirmed the reliability of the framework in Figure \ref{fig:3box1}d. Leveraging the one-to-one pixel registration, we can obtain the X, Y, and Z coordinates of the image pixels as shown in Figure \ref{fig:3box2}. Note that there were a few points that caused the contours noisy in a few locations, which was due to the numerical errors of the implemented ray tracing method and can be solved by adopting the depth-buffer technique (which is shown below). This example with the synthetic geometry shows that the proposed framework has a convincing performance to identify the 3D coordinates of each pixel in the relative simple image and geometry. The high accuracy is attributed to the outstanding performance of the key point detection scheme and the reliable optimization method design.

\begin{figure}
    \centering
    \includegraphics[width=\linewidth]{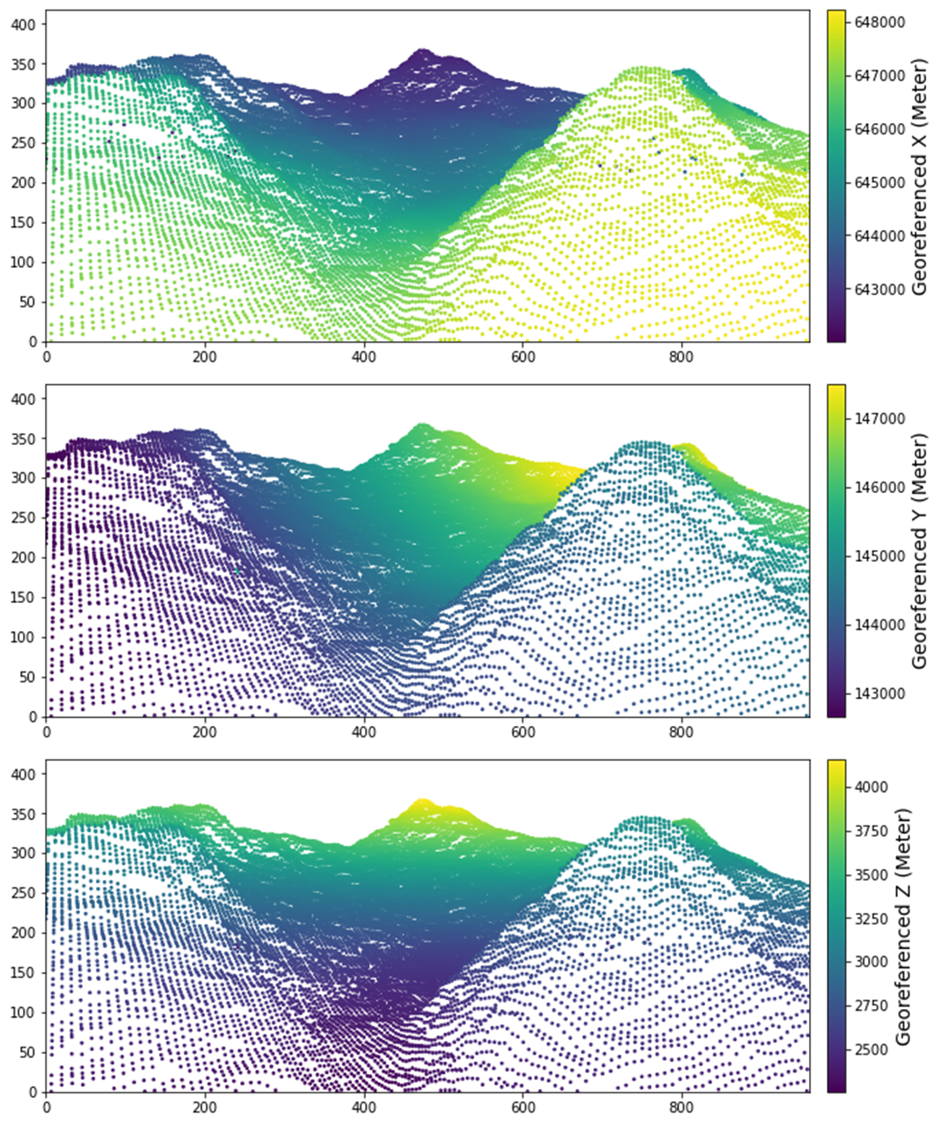}
    \caption{The obtained x, y, and z coordinates obtained from the monoplotting of the photolithograph.}
    \label{fig:Glacier2}
\end{figure}

\begin{figure}
    \centering
    \includegraphics[width=\linewidth]{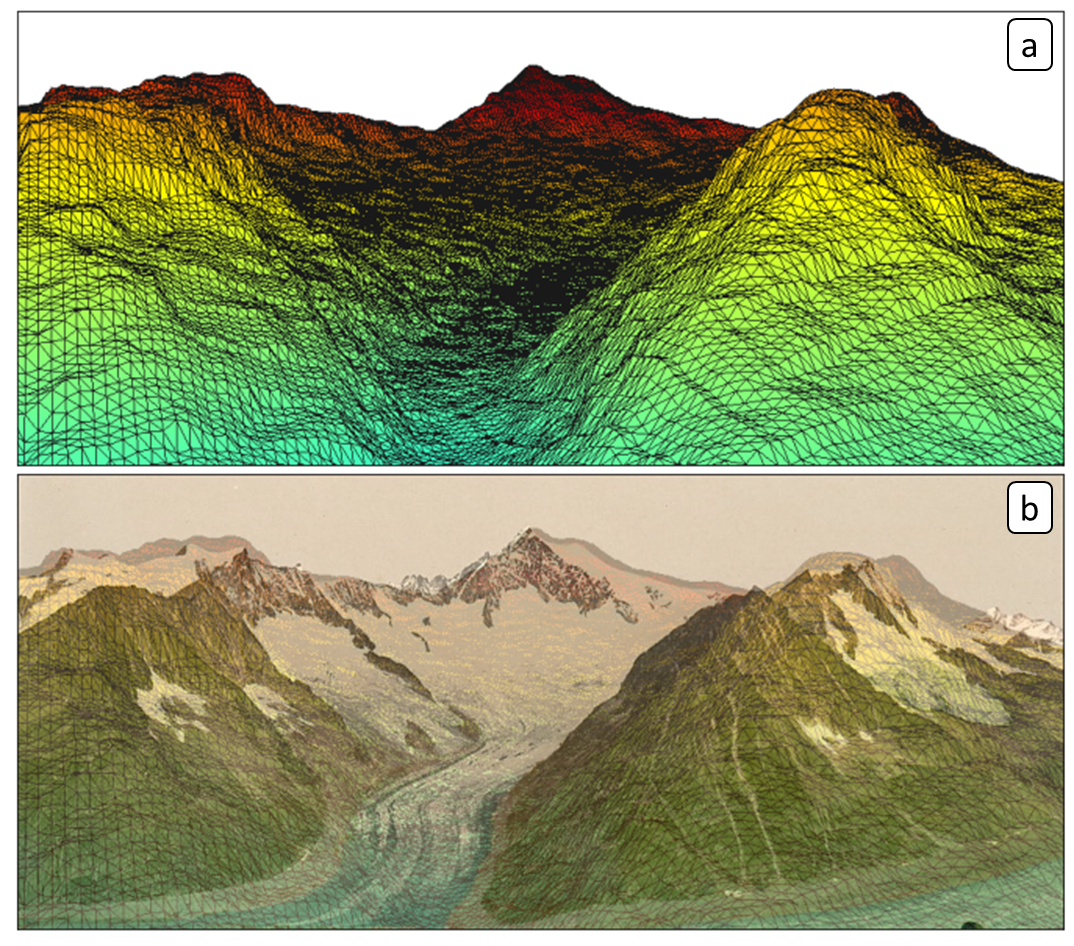}
    \caption{(a) Rendered elevation data using the obtained camera parameters; (b) An overlay of the semi-transparent photolithograph and the rendered elevation map (same as in (a)).}
    \label{fig:Glacier3}
\end{figure}

The advantage of the monoplotting technique is to connect the observation in the photography and the digital map. In the second experiment, we demonstrate that the new framework allows monoplotting of a real world example. A photolithograph of the Grand Aletsch Glacier in Switzerland generated by Photoglob Company in Zürich, Switzerland, and the Detroit Publishing Company in Michigan (Figure \ref{fig:Glacier1}) \cite{noauthor_eggishorn_1890} is processed using the proposed semi-automatic monoplotting framework. The image and regional DEM were collected from the Pic2Map QGIS plugin homepage \cite{noauthor_welcome_nodate} (Figure \ref{fig:Glacier2}). The resolution of the DEM raster is 25 meters by pixel. Applying the ORB scheme, we found a series of image GCPs can be identified in Figure \ref{fig:Glacier1}a and several DEM GCPs can be identified by the AI recommendation system shown in Figure \ref{fig:Glacier1}c. Four of the recommended DEM GCPs are selected and projected to the visual plane to compare with the image GCPs in Figure \ref{fig:Glacier1}b. Again, the optimization scheme could match the image and DEM GCPs in spite of the different numbers of them. The X, Y, Z coordinates of every pixel can then be obtained by ray tracing as shown in Figure \ref{fig:Glacier2}. The rendered DEM grid in Figure \ref{fig:Glacier3}a shows a great consistency with the photolithograph. The rendered scene is then overlapped by the photolithograph in Figure \ref{fig:Glacier3}b, in which the rendered and the original photolithograph aligned quite close. This comparison shows that the developed semi-automatic monoplotting technique could successfully identify the coordinates of every pixel in the visual data. The excellent comparison allows the analysis of glaciers: the ice distribution in the photolithograph captured in 1890 can be 3-D reconstructed using the pixel-level correspondence. This provides a unique and valuable opportunity to investigate the ancient glacier history and the movement of the glacier by comparing with more recent visual or measured data.

It is worth noting that a few gaps are identified in the numerical experiments and future improvements are planned to transform the present methodology toward a fully automatic and more robust system. First, the regularized gradient-based optimization method still requires careful design of the starting points -- some initial values resulted in local minimums. We plan to modify the regularization terms in the loss function to mitigate the problem of such local minimums. Second, adjusting the hyper-parameters of the key point detectors still needs manual tuning to be customized for different applications. To tackle this problem, deep learning key point detection models will be employed to detect key points that are most appropriate for the DEM and image GCPs. Third, in the numerical experiment, some extra key points out of the camera view could be detected, which still needed human intervention to remove. This could be improved by utilizing or assuming the view of the camera to address. Finally, in our implementations ray tracing method was employed for solving the visibility problem. Ray tracing is computationally expensive and a new version of the framework under development with switch to the depth buffer algorithm for higher reliability.

\section{Conclusion}
Photography and visual data provide rich details for documenting the real world. The traditional monoplotting method can connect visual data (especially oblique photos) and the geography to provide monocular depth estimate, image registration, and pixel georeferencing, but it cannot be scaled due to the heavy labor input loads, the requirement of users' expertise, and the difficulty of determining the camera's position and pose. 

We created a new, AI-aided georeferencing framework for pixel-level georeferencing of imagery data. The innovative use of AI transformed the traditional operation so that semi-automatic monoplotting was shown feasible. A pipeline of analyses were developed to validate the monoplotting results. To the best of authors' knowledge, this paper is the first time to 1) introduce key point detection into monoplotting, 2) apply regularized gradient-based optimization to significantly improve the accuracy of camera position and parameter determination, and 3) enable the matching of unequal numbers of GCPs from the image data and the DEM. Two numerical experiments with synthetic geometry and real world example showed that the developed framework is reliable and useful, producing satisfactory pixel-level georeferencing results. 

The new methodology described in this paper will allow pixel-level correspondence between photos and digital geographic information. The breakthrough in automatizing the operation of monoplotting will enable semi-automatic or fully automatic analysis of oblique visual data to benefit the wide researchers and industrial users, especially in processing large volume of data and providing near-real-time analysis. And finally, it will enable large-scale scaling and quick learning of monoplotting.



%



\section*{Acknowledgment}
The authors would like to acknowledge the contribution of Drs. Åsa Rennermalm and Richard Lathrope for suggestions on writing, and Dr. Yixing Yuan for fruitful discussion. In addition, they acknowledge the financial support offered by the USDOT/UTC program and Rutgers University.


\ifCLASSOPTIONcaptionsoff
  \newpage
\fi





\bibliographystyle{IEEEtran}
\bibliography{IEEEabrv}

\vfill


\end{document}